\documentclass[11pt,a4paper]{article}
\usepackage[hyperref]{acl2020}
\usepackage{times}
\usepackage{latexsym}
\usepackage{graphicx}
\usepackage{todonotes}
\usepackage{multirow}

{\ttfamily\small}

\usepackage{microtype}

\aclfinalcopy % Uncomment this line for the final submission
\title{Mitigating Gender Bias in Contextual Word Embeddings}

\author{
Navya Yarrabelly\\
nyarrabe@andrew.cmu.edu \\
\And
Vinay Damodaran\\
vdamodar@andrew.cmu.edu\\
\And
Feng-Guang Su\\
fengguas@andrew.cmu.edu \\
}

\date{}

\begin{document}
\maketitle
\begin{abstract}

Word embeddings have been shown to produce remarkable results in tackling a vast majority of NLP related tasks. Unfortunately, word embeddings also capture the stereotypical biases that are prevalent in society, affecting the predictive performance of the embeddings when used in downstream tasks. While various techniques have been proposed \cite{bolukbasi2016man, zhao2018learning} and criticized\cite{gonen2019lipstick} for static embeddings, very little work has focused on mitigating bias in contextual embeddings. In this paper, we propose a novel objective function for MLM(Masked-Language Modeling) which largely mitigates the gender bias in contextual embeddings and also preserves the performance for downstream tasks. Since previous works on measuring bias in contextual embeddings lack in normative reasoning, we also propose novel evaluation metrics that are straight-forward and aligned with our motivations in debiasing. We also propose new methods for debiasing static embeddings and provide empirical proof via extensive analysis and experiments, as to why the main source of bias in static embeddings stems from the presence of stereotypical names rather than gendered words themselves. All experiments and embeddings studied are in English, unless otherwise specified.\citep{bender2011achieving}.
\end{abstract}

\section{Introduction}
Popular word embeddings like Word2vec \cite{mikolov2013efficient}, GloVe \cite{pennington-etal-2014-glove} BERT\citep{devlin-etal-2019-bert}, ELMo\cite{peters1802deep} etc,  have been proven to reflect social biases (e.g. race and gender) that naturally occur in the data used to train them \cite{Caliskan_2017}. Main sources for these biases include disproportionate references to male entities vs female entities, discriminating racial and religious prejudices contained in the data \cite{bolukbasi2016man}.
%For example, of over 2,000 gendered pronouns in the OntoNotes test corpus, less than 25\% are feminine. WikiCoref contains only 12\% feminine pronouns.
%With the advent of popular word embeddings such as Word2vec \cite{mikolov2013efficient}, BERT\citep{devlin-etal-2019-bert}, etc, machine learning tools for language representations are being extensively deployed in many real world systems. Unfortunately, these embeddings and models can also capture real-word biases present in society and these biases tend to get amplified when used in downstream tasks. 

%This phenomenon has been recorded and researched in numerous works such as \citep{bolukbasi2016man}, \citet{De_Arteaga_2019}, and \citet{blodgett2020language}.

There has been extensive research in the recent years to mitigate these biases for gender attributes from  static embeddings\cite{bolukbasi2016man, zhao2018learning, kumar2020nurse} and there is relatively little work done on debiasing  contextual embeddings\cite{zhao2019gender, liang2020towards}. The solutions are either by using post-processing techniques to remove inherent biases present in the learnt embeddings or by trying to mitigate the bias as part of the training procedure and some  preprocessing techniques including training data augmentation using controlled perturbations to names or demographic attributes. They substantially reduce the bias in static embeddings with respect to the same definition based on the projection of a gender neutral words onto the computed gender subspace. However, experimental analysis  by \cite{gonen2019lipstick} show that the debiasing  methods for static embeddings only hide the bias and show that the bias information can be recovered from 
the representation of “gender-neutral” words. 
\par 
In this paper, we analyse the shortcomings of the existing debiasing methods for static embeddings to understand why all male and female stereotyped words  form separate clusters. We further propose ways to mitigate it and provide our experimental analysis in section 4 of the paper.

%Recently, pretrained sentence-level contextual representations from large-scale language models such as ElMo\citep{peters1802deep} and BERT have increasingly swept their way into many downstream applications. The work by \citet{zhao2019gender} looks into debiasing ElMo by augmenting training data using controlled perturbations to names or demographic attributes. Another work by \cite{liang2020towards} used a post-processing technique to debias representations from pretrained BERT for sentence -level tasks. 

In contrast to the conventional static word embedding based models, BERT and GPT \cite{devlin-etal-2019-bert} are trained on massive amounts of data for several weeks on hundreds of machines. Hence, proposed approaches for static embeddings which include complete retraining methods\cite{zhao2018learning} or data augmentation strategies \cite{dixon2018measuring, zhao2019gender} are not feasible for debasing these models. To this end, we propose continued pre-training tasks for debiasing contextual models like BERT while retaining the performance for downstream tasks. In section 3, we discuss our proposed approaches for contextualised models and present our evaluation analysis on both debiasing tasks and downstream tasks. 

%CoVe \citep{mccann2018learned} and so on. The use of word representations from large scale language models such as GPT \citep{radford2019language} and BERT\citep{devlin-etal-2019-bert}  showcased various properties pertaining to semantic similarity associations and linear translations that allow us to make popular word associations. Several other variations of word representations which move from static to contextual representations have followed like GloVe, FastText \citep{bojanowski2017enriching}, ELMo\citep{peters1802deep}, CoVe \citep{mccann2018learned} and so on. The use of word representations from large scale language models such as GPT \citep{radford2019language} and BERT\citep{devlin-etal-2019-bert} in downstream tasks have also shown considerable improvements in numerous downstream NLP tasks.

\section{Related Works}
We have extensively covered the existing methods and techniques used in order to mitigate bias in the field for both contextual and static embeddings in Appendix sections \ref{subsec:contextualEmbeddingRelatedWorks} and \ref{subsec:staticEmbeddingRelatedWorks} respectively.
\section{Contextual Word Embeddings}
% In this section, we propose new training methods for both contextual embeddings, i.e. BERT~\cite{devlin-etal-2019-bert} to mitigate gender bias. Also, we further discuss the reason why the previous works do not work well for debiasing contextual embeddings and.

\subsection{Limitations of existing methods}
\label{sec:proposed}
\begin{figure*}[h!]
\includegraphics[scale=0.48]{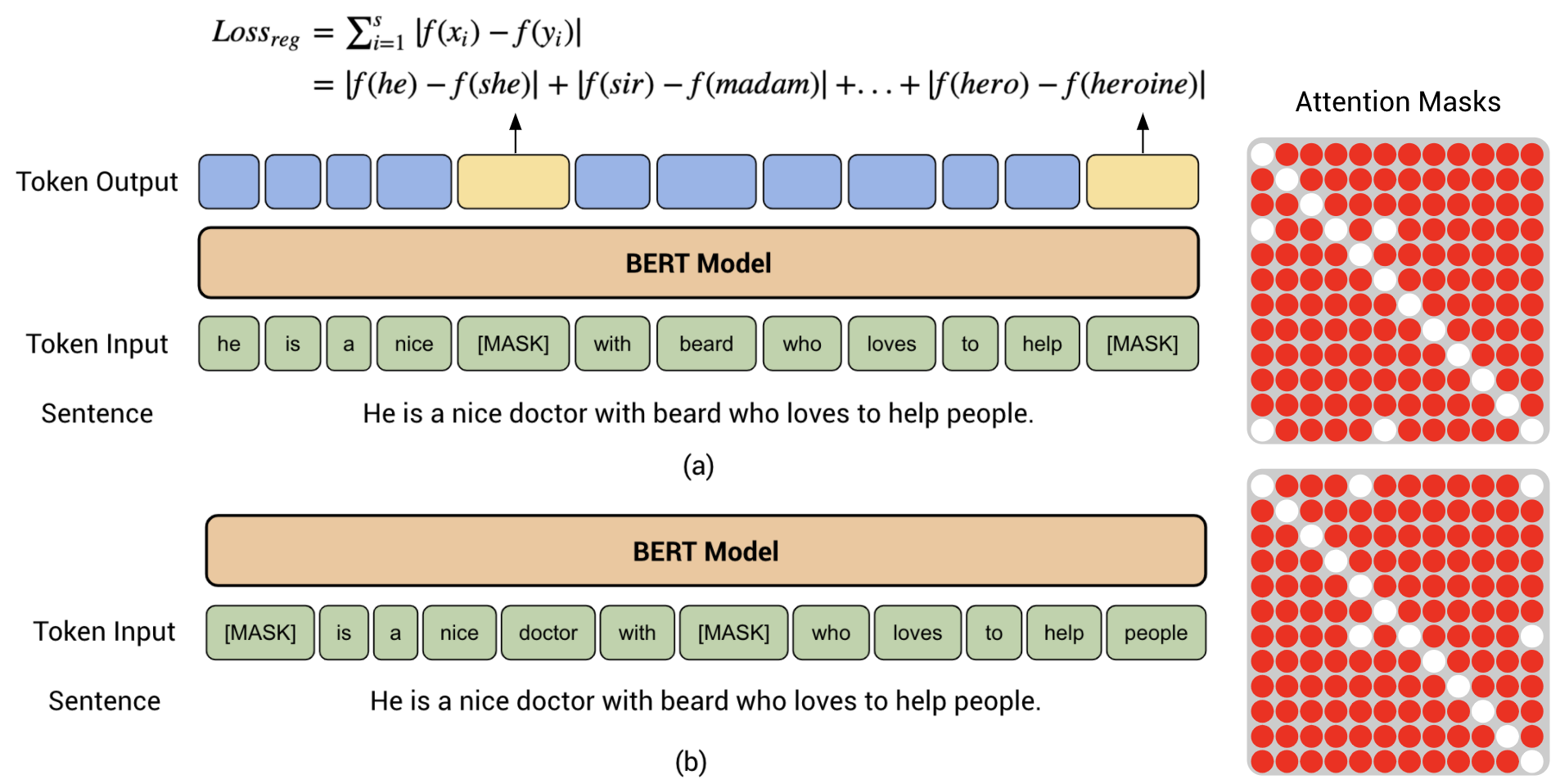}
\vspace{-8mm}
\caption{(a) We randomly mask nouns in each sentence and train the model to predict these tokens using all the other tokens of the sequence except the attribute words. (b) We randomly mask attribute words in each sentence and train the model to predict these tokens using all the other tokens of the sequence except the nouns. Note that the regularizer only applies for the case (a). The figures on the right hand side demonstrates the actual attention masks for both cases.}
\label{fig:model}
\end{figure*}

Existing studies mostly focus on identifying gender bias in static word representations such as GloVe~\cite{bolukbasi2016man}. We aim to investigate these intrinsic biases under contextualized settings by debiasing contextual embeddings.
%Unlike static word-level embeddings such as GloVe and word2vec which can be retrained on a single machine, recent works like BERT and GPT are trained on massive amounts of data and requires training for weeks. Therefore, instead of training from scratch, we choose to propose a method that can debias contextual embeddings via fine-tuning.

One of the most naive ways to debias contextual embeddings is counterfactual data augmentation (CDA). We can identify gendered pronouns/words in sentences first, replace them with words of the opposite gender, and add them to the dataset. This simple yet effective technique \citep{zhao2019gender} has proved to be very powerful in reducing gender bias in semantic similarity and coreference resolution tasks. However, there are several underlying problems which make it unsuitable to directly apply to contextualized LMs like BERT. In addition to the limitation of difficulty in training with CDA, it also eliminates the semantic information of the unpaired gender-oriented words such as bikini and beard, which can directly undermine the performance for downstream tasks.

%Firstly, recent works like BERT and GPT require massive amounts of data and weeks of training. Therefore, naive data augmentation will make the training process even harder. Second, other than debiasing gender-neutral words, CDA can also eliminate the semantic information of the unpaired gender-oriented words such as bikini and beard, which can directly undermine the performance for downstream tasks.

For the methods that are generally used and have shown to be useful for debiasing static embedding, recent works such as \citet{bolukbasi2016man}, \citet{zhao2018learning}, \citet{kumar2020nurse}, and \citet{wang2020double} may not be effective for contextual embeddings. Some of the methods mentioned above focus on post-processing by revising the word embeddings. Therefore, they cannot directly apply to the contextual embeddings. Also, even though the embeddings of gender-neutral words do not contain any gendered information through these debiasing methods such as adversarial training, there is still dependence between attribute words and target words because of the mechanism of self-attention. That is, it can still make gender-neutral words like nurse and homemaker close to each other since they have similar attribute words in context. As a result, to debias contextual embeddings like BERT, the top priority is to remove the dependence between attribute words and target words. 

\subsection{Proposed methods}
\subsubsection{Regularized MLM objective}
Here, we propose a new MLM (Masked-Language Modeling) objective function that can successfully eliminate the dependencies and meanwhile retain the model performance. Specifically, we have two strategies. We regard all nouns except the attribute words as gender-neutral words. First, we randomly mask nouns in each sentence and train the model to predict these tokens using all the other tokens of the sequence except the attribute words. On the other hand, we randomly mask attribute words in each sentence and train the model to predict these tokens using all the other tokens of the sequence except the nouns. The fig.~\ref{fig:model} clearly demonstrates the proposed training strategies and the corresponding attention masks.

To make the training process more stable and also improve the debiasing effects, we further propose a regularization method. We consider that the model should produce similar probabilities of all paired attribute words for all the gender-neutral words. Given pairs of gendered tokens $T_A = \left \{ (x_1, y_1), (x_2, y_2), ... (x_s, y_s)  \right \}$ where $x_i$ represents male tokens while $y_i$ represents the corresponding female tokens, we propose a newly-designed regularizer when calculating the cross-entropy losses of gender-neutral tokens. That is,

\begin{equation}
Loss_{reg} = \sum_{i=1}^{s} \left | f(x_i) - f(y_i) \right |, 
\end{equation}

where $f(\cdot)$ represents the prediction score of the language modeling head before softmax.

\subsubsection{Gender prediction task for CDA}
\label{BERT:genderexplicitloss}
To overcome the disproportionate number of references to male and female terms in the corpus, we propose a gender prediction task as a way to intelligently augment the data. We propose a prediction task for male, female, and neutral words in a given sentence. We use two strategies to debias the model by augmenting input data with altered gendered references and train on gender prediction task. 

\textbf{Strategy1}:  We mask out all the gendered words and PAD 40 percent of gender neutral words. We only predict the labels for gendered words ignoring the neutral words. To augment the data, we swap the labels of all the gendered words(a masked male word is set to have a target label of female and vice versa). The goal is to be able to predict the swapped gender of the words from the gender neutral words as context. By swapping genders of the masked words, we expose both male references and female references equally to gender neutral words. 

\textbf{Strategy2}: We mask all the gendered words from a sentence and predict the gender as neutral  for the gender neutral words. This helps the model to not withhold any gender information in its embedding for a neutral word. Unlike regular CDA approach used by \cite{zhao2019gender} which swaps each male word with an equivalent female word regardless of the context, we are not restricted to paired gendered words.   

\subsection{Datasets}
\label{subsec:contextualEmbeddingDatasets}
There are mainly two datasets that are used for the training and evaluation respectively. For training, we use the bookcorpus dataset~\citep{Zhu_2015_ICCV} which is also used to pretrain BERT~\cite{devlin-etal-2019-bert}. It is a rich source of both fine-grained information, how a character, an object or a scene looks like, as well as high-level semantics, what someone is thinking, feeling and how these states evolve through a story. For the preliminary evaluation, the templates are from the Winogender Schemas~\citep{rudinger-EtAl:2018:N18}, which are pairs of sentences that differ only by the gender of one pronoun in the sentence, designed to test for the presence of gender bias in automated coreference resolution systems. 

For the evaluation metric proposed by \citet{nangia-etal-2020-crows}, we test our models on WinoBias~\citep{rudinger-EtAl:2018:N18} so we can compare performance across models. Here, we utilize both dev set and test set provided by WinoBias~\footnote{https://github.com/uclanlp/corefBias}. Moreover, WinoBias has two types of sets: WinoBias-knowledge (Type1) where coreference decisions require world knowledge, and WinoBias-syntax (Type2) where answers can be resolved using syntactic information alone.

\subsection{Quantitative Analysis}
\label{sec:quantitative}

% Since there is no unified method to evaluate bias across models, we apply use different evaluation metrics to word embeddings and contextual embeddings separately.

% \subsection{Contextual Embeddings}
% \label{subsec:quantitative_contextual}

\begin{figure*}[h!]
\includegraphics[scale=0.51]{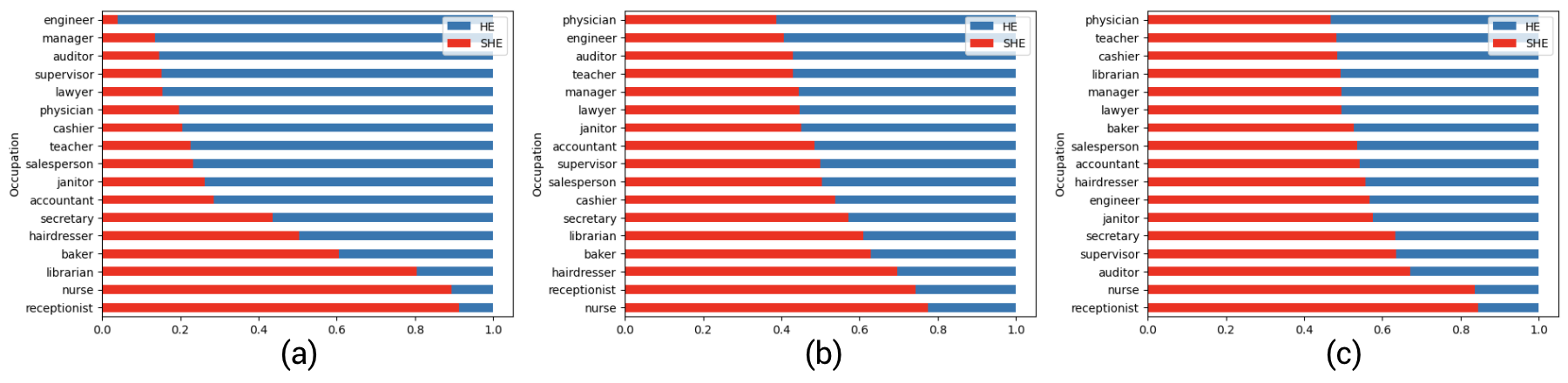}
\vspace{-8mm}
\caption{The predicted probabilities of gendered pronouns (he/she), object pronouns (him/her), or possessive adjectives (his/her) across different occupations from (a) \textbf{BERT\_based}, (b) \textbf{Proposed}, and (c) \textbf{Proposed\_reg}. The y-axis represents different occupations and x-axis represents normalized probabilities. The more identical the both probabilities are, the less bias a model obtains.}
\label{fig:preliminary}
\end{figure*}

\begin{figure*}[h!]
\includegraphics[scale=0.46]{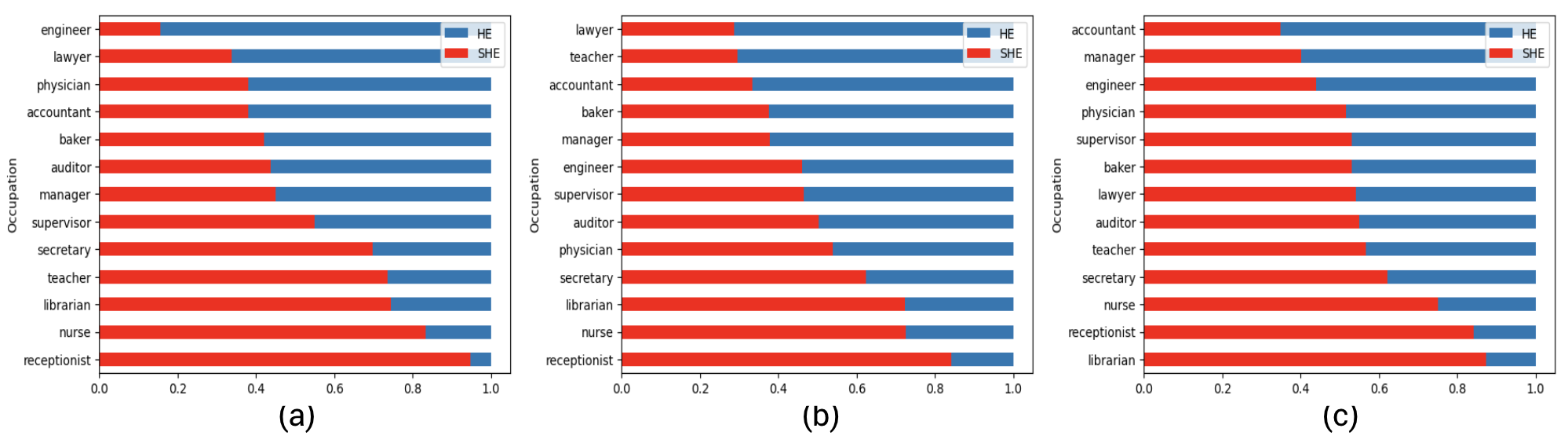}
\vspace{-8mm}
\caption{The predicted probabilities of occupations based on similar contexts with different gendered pronouns (he/she), object pronouns (him/her), or possessive adjectives (his/her) across different occupations from (a) \textbf{BERT\_based}, (b) \textbf{Proposed}, and (c) \textbf{Proposed\_reg}. The y-axis represents different occupations and x-axis represents normalized probabilities. The more identical the both probabilities are, the less bias a model obtains.}
\label{fig:preliminary_job}
\end{figure*}

Here, we will report different evaluation metrics on 4 different models, including \textbf{BERT\_based}, \textbf{Proposed}, \textbf{Proposed\_reg}, \textbf{Proposed\_aug}. They symbolize the BERT pretrained model, the proposed post-processing method, the proposed method with regularization and the proposed method of data augmentation with gender prediction task respectively.

\subsubsection{Preliminary Analysis}
\label{subsubsec:preliminary}

To measure the underlying gender bias in the contextual embeddings, the most straight-forward way is to predict the masked words (attribute words and target words) based on similar context. We select various occupations from U.S. Bureau of Labor Statistics \footnote{https://www.bls.gov/cps/cpsaat11.htm} that are socially gender biased for the target words. Then based on the occupations, we sampled several templates from the Winogender Schemas~\citep{rudinger-EtAl:2018:N18} \footnote{https://github.com/rudinger/winogender-schemas} and use them to evaluate the model. To achieve the goal, we design 2 brand-new evaluation tasks. 

First, given a template like "The engineer informed the client that he/she would need more time to complete the project.", we can mask the pronouns (he/she) and fill the mask by predicting all the possible words. Therefore, the larger the difference of the probabilities of pronouns (he/she) is, the more bias that a model obtains. The results shown in the fig.~\ref{fig:preliminary} demonstrate that \textbf{proposed} and \textbf{proposed\_reg} successfully debias the contextual embeddings for most of the words since the probabilities of pronouns (he/she) are rather closer. Specifically, we can find that \textbf{proposed\_reg} has better performance since the probability of each pronoun is closed to 0.5 for most of the occupations, which means that they are absolutely gender-neutral. We observed that our augmentation approach \textbf{proposed\_aug} although encoded information equally w.r.t the gender, has assigned very low probabilities for he/she tokens when predicting for \textit{MASK is a doctor}. We suspect that the debiasing procedure with gender prediction task has stripped all the information encoded to predict gendered pronouns from a gender-neutral word. 

Similarly, we can mask the occupations (i.e. engineer) in the template and fill the mask by predicting all the possible words based on similar context with different pronouns (he/she). If the probability of the target words (i.e. engineer) is larger when we use "he" rather than "she" in the context, it means that the model regarded the target words as male-related words instead of gender-neutral words. So as shown in Fig.~\ref{fig:preliminary_job}, \textbf{proposed} and \textbf{proposed\_reg} again outperforms the baseline model \textbf{BERT\_based}. Also for most of the words, the probabilities remain the same based on different gendered pronouns using the \textbf{proposed\_reg} method. Note that this evaluation metric is much harder than the previous one since it is more difficult for a model to predict the specific occupations based on the context. Therefore, we only report the results where the target occupations appear in the top 30000 predicted words. 

\begin{table}[h!]
\small
\centering
\begin{tabular}{|c|c|c|c|c|}
\hline
\textbf{Dataset} & \multicolumn{2}{|c|}{Type1} & \multicolumn{2}{|c|}{Type2} \\ \hline
\textbf{Model} & dev & test & dev & test \\ \hline
\textbf{BERT\_based}   & 56.57 & 53.54 & 60.10 & 60.61 \\ \hline
\textbf{Proposed}      & \textbf{53.28} & 50.76 & 57.07 & \textbf{52.27} \\ \hline
\textbf{Proposed\_reg} & 55.05 & 51.52 & \textbf{54.04} & 54.04 \\ \hline
\textbf{Proposed\_aug} & \textbf{46.97} & \textbf{50.25} & 54.28 & 52.53 \\ \hline
\end{tabular}
\vspace{-3mm}
\caption{The stereotype score proposed by \citet{nangia-etal-2020-crows} on WinoBias-knowledge (Type1) and syntax (Type2)~\citep{rudinger-EtAl:2018:N18}. Higher numbers indicate higher model bias. }
\label{table:stereotypeScore}
\end{table}

\subsubsection{Evaluation Analysis}
\label{subsubsec:evaluation}
For a sentence $S$, let $U = {u_0, . . . , u_l}$ be the unmodified tokens, and $M = {m_0, . . . , m_n}$ be the modified tokens, so $(S = U \cup M)$. Since we are focusing on mitigating gender bias in BERT models, we utilize WinoBias~\citep{rudinger-EtAl:2018:N18} dataset, which contains templates where the modifications in $M$ are the replacements of attribute words with their corresponding words of the opposite gender. So, the stereotype score proposed by \citet{nangia-etal-2020-crows} mainly approximate the probabilities $p(U|M, \theta)$ by adapting pseudo log-likehood MLM scoring generated by BERT models. The larger the numbers are, the higher the gender stereotype that a model obtains. 

As shown in the table~\ref{table:stereotypeScore}, our proposed methods outperform the \textbf{BERT\_based} model on all the datasets. Specifically, our proposed methods perform favorably by a large margin on the Type2 dataset. We believe the reason that the proposed methods perform better is that the answers can be resolved using syntactic information alone on the Type2 dataset, which is more aligned with our training motivations. 

\subsubsection{Sentence Embedding Association Test (SEAT)}
Sentence Embedding Association Test (SEAT) extends the logic  of WEAT\cite{Caliskan_2017} for computing bias in contextual word embeddings. The SEAT method compares set of target concepts (e.g. male and female words) denoted as X and Y (each of equal size N), with a set of attributes to measure bias over social attributes and roles (e.g. career/family words) denoted as A and B. The degree of bias for each target set X and Y is denoted by $S(X,Y,A,B)$ based on difference in similarities between target word and attributed word embeddings across X and Y. We compute the p-value for the permutation test, which is defined as
the probability that a random even partition $X_i$, $Y_i$ of $X U Y$ satisfies $P[s(X_i, Y_i, A, B) > s(X, Y, A, B)]$. When p-value is less than 0.05, we reject the null hypothesis that the observed bias effects are only due to extreme sampling of data. Hence, SEAT can meaningfully conclude the presence of bias only if the p-value is less than 0.05. 
\par We compare the SEAT results of our debiasing method with Sent-Debias\cite{liang2020towards} in table\ref{table:SEAT_CLM}. The original BERT\cite{devlin-etal-2019-bert} approach shows significant bias on most of the Caliskan test sets. Sent-Debias\cite{liang2020towards} primarily  based their evaluation on SEAT metric and the SEAT test results for their approach are not statistically significant and hence cannot conclusively prove the presence/reduction of  bias. The SEAT for debiased BERT fails to find any
statistically significant biases at p-value $\leq$ 0.05 for 4/6 of the test-sets. This
implies that SEAT is not an effective measure
for bias in BERT embeddings and is shown to have erratic effects  when  additional context information is added to the templates(shown in Assignment 3).  
\subsubsection{Downstream Task Evaluation}
To ensure that the debiasing procedure does not deteriorate performance on downstream semantic tasks, we  evaluate the performance of our debiased BERT by finetuning it on Stanford Sentiment Treebank (SST-2) for sentiment classification \cite{socher2013recursive} and Corpus of Linguistic Acceptability (CoLA) for grammatical acceptability judgment \cite{warstadt2019neural} , Question
Natural Language Inference (QNLI) \cite{wang2018glue}. We also compare our results with that from Sent-Debias \cite{liang2020towards}. 
Table~\ref{table:COLASST} shows the results obtained on the finetuning tasks. Our model achieved a 0.8\% gain over the original BERT model on SST-2 task but there is about 3\% drop in accuracy  on the CoLA task, achieved performance on the QNLI  tasks. We hypothesise that the drop in performance on CoLA task due to the fact that it tests for low level syntactic structure information such verb usage, tenses etc. The dataset also contains several proper noun references such as \textit{Harry coughed himself into a fit} and our model cannot distinguish between male names and female names at training time and hence can encode inconsistent gender aspects. 
SENT-DEBIAS has a 1-2\% drop in performance on all tasks owing to the debiasing. Our debiasing training procedure thus retained the semantic concepts associated with the words while successfully stripping the gender information encoded through biases. 

\begin{table}
\small
\centering
\caption{ Effect of debasing approaches on both
single sentence (BERT on SST-2, CoLA) and double sentence (BERT on QNLI)downstream tasks. The performance  of debiased BERT representations on downstream tasks is  retained on QNLI and SST-2 tasks.}
\vspace{-3mm}
\begin{tabular}{|c|c|c|c|}
\hline
\textbf{Test} & SST-2 & CoLA & QNLI \\ \hline
\textbf{BERT} & 92.6 & 57.6 &  91.3 \\ \hline
\textbf{Sent-Debias} & 89.9 & 54.7 & 90.6 \\ \hline
\textbf{Proposed\_reg (ours)} & \textbf{93.4} & 54.08 & \textbf{91.3} \\ \hline
\textbf{Proposed\_aug (ours)} & 92.08 & \textbf{54.75} & 90.48 \\ \hline
\end{tabular}
\label{table:COLASST}
\end{table}

\begin{table*}
\small
\centering
\begin{tabular}{|c|c|c|c|}
\hline
\textbf{Test Set} & \textbf{BERT} & \textbf{Debiased-bert ours} & \textbf {SENT-DEBIAS BERT}  \\ \hline
C6: M/F Names, Career/Family & +0.477* & \textbf{0.429}* (0.008) & -0.0897 (0.69)  \\ \hline
C6b: M/F Terms, Career/Family & +0.108 & 0.283* (0.03) &  -0.434 (0.997)\\ \hline
C7: M/F Names, Math/Arts & +0.253* & -0.68 (0.99)  &+0.194 (0.12)\\ \hline
C7b: M/F Terms, Math/Arts & +0.254* & \textbf{-0.0189*} (0.04) & +0.193 (0.12) \ \\ \hline
C8: M/F Names, Science/Arts & +0.399* & -0.154  &   -0.071 (0.64) \\ \hline
C8b: M/F Terms, Science/Arts & +0.636* & 0.72 (0.63) & \textbf{+0.54}* (0.002) \ \\ \hline
\end{tabular}
\vspace{-3.5mm}
\caption{Debiasing results on BERT sentence representations. First six rows measure binary SEAT effect sizes for sentence-level tests. SEAT scores closer to 0 represent lower bias. Numbers in brackets indicate the  p-values for the test. * denotes that p-value is less than 0.05 and the test result is statistically significant.}
\label{table:SEAT_CLM}
\end{table*}

\subsection{Qualitative Analysis}
\label{sec:qualitative}

\begin{figure}[h]
\centering
\includegraphics[scale=0.30]{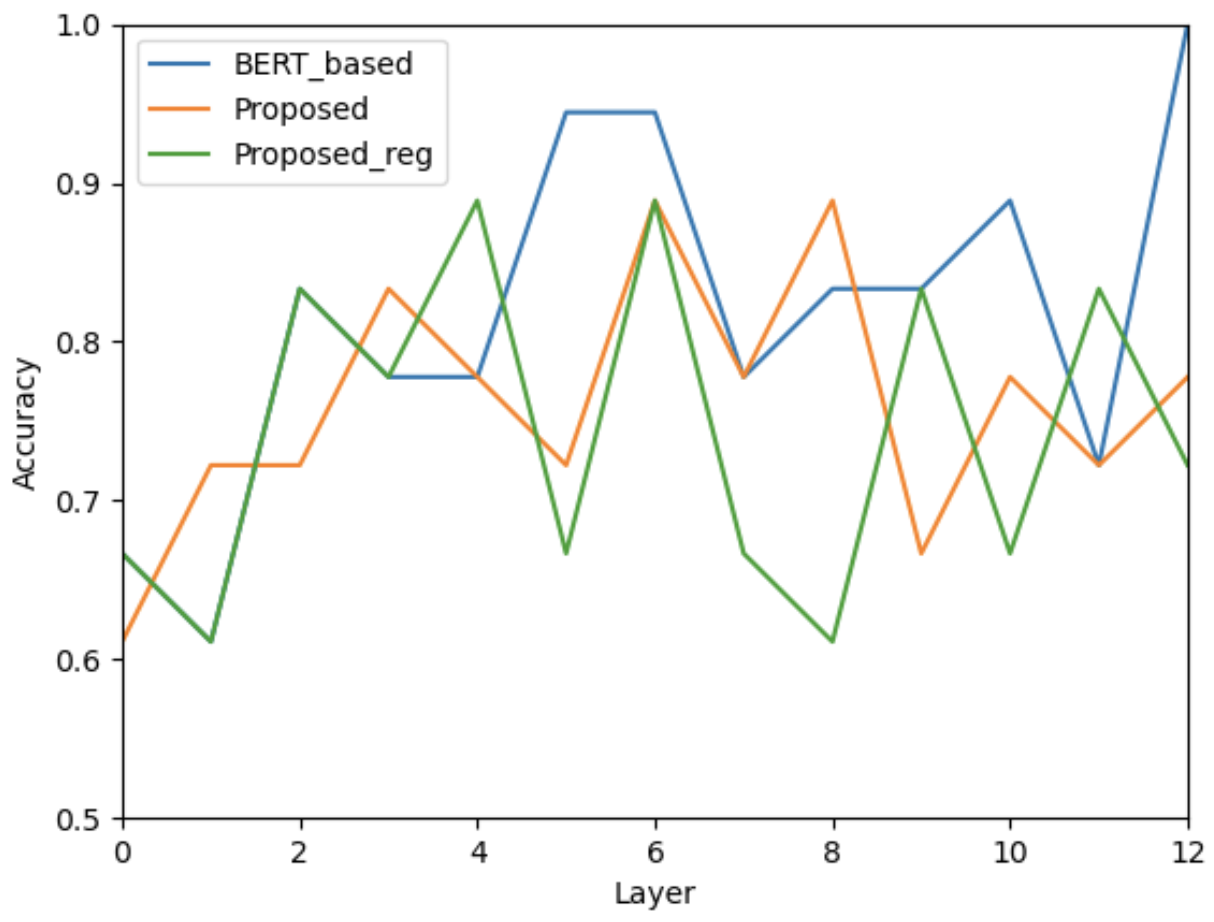}
\vspace{-8mm}
\caption{Classification accuracy for each layer in BERT. The classification model here is a linear classifier with early stopping. Better accuracy represents more gender bias encoded in a model.}
\label{fig:lrAnalysis}
\end{figure}

\subsubsection{Layer Analysis}
\label{subsec:layerAnalysis}
To analyze when the gender bias is encoded in BERT, we collect the embeddings from each layer and the labels from U.S. Bureau of Labor Statistics. Specifically, the embeddings are the features of gendered pronouns (he/she),  object pronouns (him/her),  or possessive adjectives (his/her) that refer to different occupations in the templates from Winogender Schemas~\citep{rudinger-EtAl:2018:N18}. As shown in the fig.~\ref{fig:lrAnalysis}, \textbf{BERT\_based} model incorporates gender bias in the fifth layer while the accuracies of \textbf{Proposed} and \textbf{Proposed\_reg} fluctuate with layers. We suspect the reason why the accuracies tend to be high is that the dimension is too large for binary classification, and therefore we apply a linear classifier with early stopping so that we can observe the difference between layers and models.

\section{Static Word Embeddings}
\subsection{Limitations of existing methods}
Various experiments for analysis \citep{gonen2019lipstick} have pointed out that existing post-processing debiasing techniques do nothing more than superficially remove the presence of bias. One of the key observations \citep{gonen2019lipstick} is that stereotypically male and female gendered professions still cluster together even after post-hoc debiasing techniques. While the direct bias with the words woman/man or she/he have been masked, the embeddings still contain information that causes an indirect association with words that are stereotypically of the same gender.  This has proven to be the case for all modern debiasing techniques including \citet{bolukbasi2016man}, \citet{kumar2020nurse} and \citet{wang2020double}. \citet{wang2020double} postulates that there might be factors other than gender that can cause this clustering. We provide a series of experiments to understand where this bias stems from and try to find ways to mitigate the aftereffects of this inadvertent clustering.

\subsection{Proposed methods}
\subsubsection{Representing gendered words as semantic concepts}
In order to understand whether the bias is due to the context of gendered words co-occurring with the gender-neutral professions, we take a standard text corpus and create a gender-neutral version of this corpus. We take a predefined list of male and female gendered words \citep{bolukbasi2016man} and concatenate the pairs into a single word. This single word is then treated as a single semantic concept and replaced throughout the corpus to represent both gendered versions of the word.

The intuition is to analyze whether presence of gendered words in a corpus has an impact in imbibing the stereotypical correlations between gender-neutral professions. If the presence of words like he/she are actual cues that influence the embedding space and bias gender-neutral professions such as nurse, doctor, etc., then training the embeddings from scratch should yield results which would be less biased and more semantic in nature.

\subsubsection{Explicit Gender Encoding}
We introduce a new objective function to the existing CBOW training curriculum which acts a regularizer by explicitly teaching the model to differentiate between gendered and gender-neutral words. During training, we take the embeddings of the center words and pass them through a dense layer which predicts probabilities across 3 categories of possible gender for each word (male, female, and gender-neutral). This is similar to the approach in \ref{BERT:genderexplicitloss} where we penalize the model if it wrongly identifies the gender of a word based on it's embedding. We use the same pre-defined word list mentioned in the previous approach to annotate male and female words, and all other words are given a label as gender-neutral. The intuition behind the regularizer is to train the model to encode the gender of a word into its embedding so that it can distinguish between words that have an inherent gender (such as actor, saleswoman) versus words like nurse which is stereotypically considered female but has no real inherent gender. 

\subsubsection{Masking stereotypical named entities}
In \citet{dev2019attenuating}, stereotypical gender name pairs help determine the gender direction component via PCA as effective as the fundamental set of 10 definitional pairs provided by \citet{bolukbasi2016man} to estimate and identify the gender subspace. We further postulate that the presence of such stereotypical word names is what cause embeddings like hairdresser and nanny to cluster together despite not having common co-occurring unigrams that occur in the same context. In order to test this hypothesis, we curate a new corpus that masks out the most stereotypical and popular set of male and female names with an entity mask token, thereby preventing any spurious correlations due to the presence of such named entities.

\subsection{Datasets and Experimental Setup}
Due to the lack of compute resources, we train all embeddings on the Fil9 dataset\footnote{http://mattmahoney.net/dc/textdata.html}. This leads to embeddings of slightly lower quality than what is available as the off-the-shelf versions of Word2Vec (which are trained on substantially larger amounts of data). However, for fair comparison, we evaluate all debiasing methods and techniques on embeddings generated with the same dataset and hence expect all results to be comparable.
We use a standard CBOW model \citep{mikolov2013efficient} with batch size 4096, window size 5 and trained for 5 epochs. We use the data available from the Social Security Administration\footnote{https://www.ssa.gov/oact/babynames/limits.html} to make a list of stereotypical male and female names that have occurred more than 10000 times since 1880.  
We use this curated list to create a new NER-Masked dataset that replaces all named entities with an entity token. The loss used for explicit gender encoding uses a constant regularizing factor of 0.5.

\subsection{Quantitative analysis}
\subsubsection{Cluster accuracy for the most stereotyped professions}
\label{subsubsec:clusterAcc}
In order to test the efficacy of our proposed methods, we evaluate the debiased embeddings by conducting the gender-profession clustering experiment proposed by \citet{gonen2019lipstick}.

We evaluate our proposed methods and the baselines by calculating the classification accuracy of the 50 most stereotypically biased male and female professions divided into 2 clusters\footnote{Due to the small size of our dataset, some professions are not present in the embedding vocabulary and finally we end up with a list of 91 professions (46 male/45 female)}. The list of biased professions themselves have been taken from \citet{bolukbasi2016man} who in turn have collected the data through mechanical turkers and crowd-sourcing.
An embedding that is heavily biased will score a higher accuracy while clustering the professions and an unbiased system, would score a classification accuracy closer to 50\%. The accuracy is measured by determining the cluster-stereotype grouping that gives the highest accuracy. The lower the clustering accuracy is, the less likely the embeddings conform to gender stereotypes in professions. The results for the original embeddings and the baselines are shown in table \ref{table:clusterAccuracy}. 

\begin{table}[h!]
\small
\begin{tabular}{|c|c|}
\hline
\textbf{Embedding}         & \textbf{Accuracy} 
\\ \hline
Original \citep{mikolov2013efficient}                        & 0.5802 $\pm$ 0.0282    \\ \hline
Hard Debias \citep{bolukbasi2016man}                    & 0.5832 $\pm$ 0.0205                       \\ \hline
Double Hard Debias \citep{wang2020double}              & 0.5714 $\pm$ 0.0256                      \\ \hline
RAN Debias \citep{kumar2020nurse}                      & 0.5538 $\pm$ 0.0368                       \\ \hline
Gender-neutral (Ours)           & 0.5970 $\pm$ 0.0169                       \\ \hline
Explicit Gender Encoding (Ours) & 0.5743 $\pm$ 0.0106                       \\ \hline
NER-Masked  (Ours)              & 0.5839 $\pm$ 0.0356                       \\ \hline
NER-M + EGE (Ours)              & \textbf{0.5157 $\pm$ 0.0050}              \\ \hline
\end{tabular}
\vspace{-3mm}
\caption{The clustering accuracy of gender-neutral professions with stereotypically biased professions (averaged across 15 runs) for various debiasing techniques. Lower accuracy indicates less bias. An exhaustive list of the cluster labels for the best performing approach is available in Appendix \ref{sec:appendix:clusterList}.}
\label{table:clusterAccuracy}
\end{table}

We observe that apart from \citet{kumar2020nurse}, other methods do not outperform the original embeddings which substantiates the claims by \citet{gonen2019lipstick} of how remnants of bias remain in the so-called "debiased" embeddings.
Our proposed approaches also fail to successfully disentangle the gender bias associated with these professions;however interestingly, masking the stereotypical male/female names succeeds slightly better than removing gendered words completely, which might suggest that names are perhaps more potent and silent carriers of bias than actual gendered words themselves. 

A combination of both NER-Masking (NER-M) and Explicit Gender Encoding (EGE) seem to have a profound effect on reducing the correlations between gender-stereotypical groups of professions. This bodes well with the idea of how gender can be encoded into an embedding and we can remove any traces of residual bias by identifying and nullifying the carriers of bias between these groups, which are likely to be names. Further empirical proof of how names act as the silent carriers of gender bias is explored in sec~\ref{sec:staticqualitative}. 

\subsubsection{SemBias scores}
First introduced in \citet{zhao2018learning}, we evaluate our debiased embeddings on the SemBias dataset to understand how well they perform when it comes to identifying both gender-definition word pairs (predicting the equivalent word with respect to he-she as an analogy) and stereotypical gender pairs. A high quality debiased embedding would score high for the definitional word pairs and low for stereotypical pairs.
We observe that the proposed model does indeed improve in quality and reduces compared to the original embedding which reinforces the intuiton behind the proposed techniques. A summary of the results is provided in Table \ref{table:sembias}.

We notice that RANDebias and Hard Debias both perform significantly better than the proposed approaches, which we attribute due to the nature of how SemBias scores are calculated. Both methods directly take into account the gender direction through the difference of the gendered pronouns (he - she) and inculcate methods directly optimizing the effect of this direction on biased words. However, as shown in Section \ref{subsubsec:clusterAcc} this does not help these techniques to reduce correlations between stereotypical instances of gender-neutral words whereas our method showcases improvement in mitigating bias on a multi-faceted level. This metric relies on the singular notion that the bias direction can be captured by the difference in gendered pronouns. It is due to similar reasons that we avoid comparison of bias with the GIPE score proposed by \citet{kumar2020nurse} as the metric penalizes neighbours that may be stereotypical without regard to its semantic nature.

\begin{table}[]
\small
\begin{tabular}{|c|c|l|}
\hline
\multirow{2}{*}{\textbf{Embedding}} & \multicolumn{2}{c|}{\textbf{Sembias}}             \\ \cline{2-3} 
                                         & D      & \multicolumn{1}{c|}{S} \\ \hline
Original \citep{mikolov2013efficient}                                 & 0.7225          & 0.175                           \\ \hline
Hard Debias \citep{bolukbasi2016man}                              & 0.7325          & 0.08                            \\ \hline
Double Hard Debias \citep{wang2020double}                      & 0.3225          & 0.25                            \\ \hline
RAN Debias \citep{kumar2020nurse}                              & \textbf{0.8875} & \textbf{0.0425}                 \\ \hline
% Explicit Gender Encoding (Ours)          & 0.605           & 0.2125                          \\ \hline
% NER-Masked  (Ours)                       & 0.685           & 0.1475                          \\ \hline
NER-M + EGE (Ours)                       & 0.74            & 0.125                           \\ \hline
\end{tabular}
\vspace{-3mm}
\caption{Results of analogy accuracy for definitional (D) and stereotypical (S) pairwise analogies in SemBias. Higher and lower values are preferred for definitional and stereotypical word pairs respectively.}
\label{table:sembias}
\end{table}

\begin{figure*}[h!]
\centering
\includegraphics[scale=0.47]{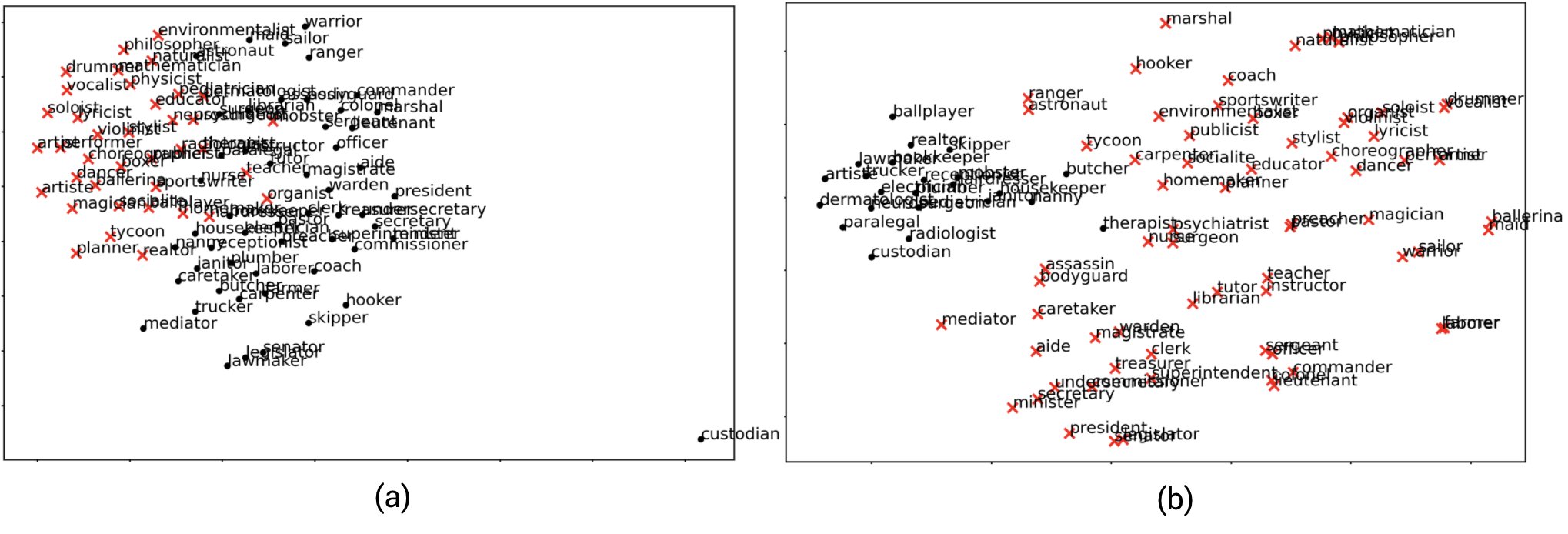}
\vspace{-9mm}
\caption{The gender clustering visualization for (a) \textbf{Original Word2vec}, (b) \textbf{NER-M + EGE Word2Vec}. We can observe clear semantic groupings between the professions (specifically clusters around the words ``teacher", ``nurse", ``commander", ``mathematician", ``pastor", ``assassin" ``lyricist" to name a few).  Visualizations for other baselines and approaches are provided in Appendix \ref{appendix:genderCluster}}
\label{fig:cluster1}
\end{figure*}

\subsection{Qualitative analysis}
\label{sec:staticqualitative}

\subsubsection{Gender clustering for professions}
We plot the TSNE \citep{van2008visualizing} visualizations in fig.~\ref{fig:cluster1} for the embeddings of the professions and find that the vectors appear to form small niches or ``mini-cluster" based on the semantic relatedness of their professions. These clusters are much more well pronounced and spread out with well demarcated boundaries as compared to original visualizations of the same embeddings. We also find a number of interesting clusters that seem to represent the various fields. A few notable observations are how words like `nurse' shift to different clusters after debiasing and words like `hairdresser' and `nanny' move further apart in the semantic space. However, words like `receptionist' and `bookkeeper' seem to maintain their relative semantic spaces indicating that only the debiasing technique proposed is able to differentiate between correlations based on gender and those between intersection of fields. A less congested visualization of this phenomenon is provided in Appendix \ref{appendix:less_annotate}.

\subsubsection{Neighbour analysis for words}
Another evaluation metric proposed by \cite{gonen2019lipstick} suggests that another indirect form of measuring bias is by measuring the Bias-by-projection which correlates to bias-by-neighbours. We use this qualitative evaluation technique to observe shift in neighbouring words for some of the more biased stereotypical professions such as nurse, nanny and observe how the behavior of its neighbours change with the proposed methods. 

We provide a word cloud visualization for the neighbours of the word `nanny' and `hairdresser' in Appendix \ref{appendix:hairdresser_nanny}. We see that the EGE technique is successful in disentangling female associated words from an occupation; however, female stereotypical names still cluster closely suggesting that the source of bias in professions is due to the presence of similarly clustering names but not the gendered pronouns/words themselves. This justifies the intuition behind the NER-M technique as completely eliminating the bias manifested in these embeddings.  We also observe the reverse of this process in Appendix \ref{sec:appendix:molly} where we see how female gendered professions move away from female stereotypical names and rather cluster together with similar albeit names of the same stereotype.  This successful disentanglement of gender-enforced professions to names can be considered as one of the primitive starting steps to build a more inclusive set of embeddings for the wider non-binary gender community.
We also visualize and analyze the change in patterns and semantic associations of neighbours for the word `nurse' across baselines and our proposed approaches in Appendix \ref{fig:nursebig}.

\subsection{Limitations and Future Work}
Due to the limited dataset that we train the embeddings on, we do not fully capture a vocabulary big enough to evaluate and compare the quality of the debiased word embeddings with respect to various word analogy tasks. The NER-M + EGE method although it correctly identifies the source from where bias stems from, does also fails to provide debiased embeddings for the names removed. Modifying the proposed approach to account for this is one of the potential directions to extend this future work.

We also conduct a small scale analysis on whether the proposed approach is successful in mitigating the bias in common adjectives such as the word `dirty'. While our method does not succeed in disentangling the stereotypical associations completely, we do observe a subtle improvement in the neighbouring space although not completely unbiased (provided in Appendix \ref{appendix:dirty}\footnote{Note that these visualizations contain explicit content}). Further analysis needs to be conducted to understand the source of bias that is propagated through adjectives.

\section{Conclusion}
We successfully introduce and analyze new novel training curriculum and post-processing techniques to tackle gender bias in both contextual and static word embeddings. We show that our models successfully mitigate gender bias in BERT by measuring the bias in models and meanwhile maintain comparable performance for the downstream tasks. For static embeddings, we show results that showcase great promise in successfully removing the gender ingrained in stereotypical professions. 

The techniques proposed include approaches that are universal in nature and can be incorporated into any model architecture/dataset, making them model-invariant and dataset-agnostic. We also do an in-depth analysis to understand the source of bias in static word embeddings and provide various experimental results and empirical analysis to conclude that more than gendered words themselves, names act as the silent carriers of gender bias. Completely removing bias is still an unsolved task and most modern techniques act as "one-trick ponies" rather than a "silver bullet".But perhaps, the first step into changing that, is to understand why and we hope this report proves to be a positive step towards that direction.

% We successfully introduce new novel techniques
% We have introduced a post-processing (fine-tuning) method to mitigate gender bias in contextual embeddings. Also, we further proposed a regularizer to stabilize training. Since we mainly focus on revising the MLM objective function, the proposed method can directly be applied to various model such as XLNet~\citep{DBLP:journals/corr/abs-1906-08237} or ERNIE~\citep{DBLP:journals/corr/abs-1905-07129}. In this paper, 

\bibliography{anthology,acl2020}

\begin{thebibliography}{28}
\expandafter\ifx\csname natexlab\endcsname\relax\def\natexlab#1{#1}\fi

\bibitem[{Bender(2011)}]{bender2011achieving}
Emily~M Bender. 2011.
\newblock On achieving and evaluating language-independence in nlp.
\newblock \emph{Linguistic Issues in Language Technology}, 6(3):1--26.

\bibitem[{Bolukbasi et~al.(2016)Bolukbasi, Chang, Zou, Saligrama, and
  Kalai}]{bolukbasi2016man}
Tolga Bolukbasi, Kai-Wei Chang, James~Y Zou, Venkatesh Saligrama, and Adam~T
  Kalai. 2016.
\newblock Man is to computer programmer as woman is to homemaker? debiasing
  word embeddings.
\newblock \emph{Advances in neural information processing systems},
  29:4349--4357.

\bibitem[{Caliskan et~al.(2017)Caliskan, Bryson, and Narayanan}]{Caliskan_2017}
Aylin Caliskan, Joanna~J. Bryson, and Arvind Narayanan. 2017.
\newblock \href {https://doi.org/10.1126/science.aal4230} {Semantics derived
  automatically from language corpora contain human-like biases}.
\newblock \emph{Science}, 356(6334):183–186.

\bibitem[{Dev and Phillips(2019)}]{dev2019attenuating}
Sunipa Dev and Jeff Phillips. 2019.
\newblock \href {http://arxiv.org/abs/1901.07656} {Attenuating bias in word
  vectors}.

\bibitem[{Devlin et~al.(2019)Devlin, Chang, Lee, and
  Toutanova}]{devlin-etal-2019-bert}
Jacob Devlin, Ming-Wei Chang, Kenton Lee, and Kristina Toutanova. 2019.
\newblock \href {https://doi.org/10.18653/v1/N19-1423} {{BERT}: Pre-training of
  deep bidirectional transformers for language understanding}.
\newblock In \emph{Proceedings of the 2019 Conference of the North {A}merican
  Chapter of the Association for Computational Linguistics: Human Language
  Technologies, Volume 1 (Long and Short Papers)}, pages 4171--4186,
  Minneapolis, Minnesota. Association for Computational Linguistics.

\bibitem[{Dixon et~al.(2018)Dixon, Li, Sorensen, Thain, and
  Vasserman}]{dixon2018measuring}
Lucas Dixon, John Li, Jeffrey Sorensen, Nithum Thain, and Lucy Vasserman. 2018.
\newblock Measuring and mitigating unintended bias in text classification.
\newblock In \emph{Proceedings of the 2018 AAAI/ACM Conference on AI, Ethics,
  and Society}, pages 67--73.

\bibitem[{Gonen and Goldberg(2019)}]{gonen2019lipstick}
Hila Gonen and Yoav Goldberg. 2019.
\newblock Lipstick on a pig: Debiasing methods cover up systematic gender
  biases in word embeddings but do not remove them.
\newblock \emph{arXiv preprint arXiv:1903.03862}.

\bibitem[{Hovy et~al.(2006)Hovy, Marcus, Palmer, Ramshaw, and
  Weischedel}]{10.5555/1614049.1614064}
Eduard Hovy, Mitchell Marcus, Martha Palmer, Lance Ramshaw, and Ralph
  Weischedel. 2006.
\newblock Ontonotes: The 90% solution.
\newblock In \emph{Proceedings of the Human Language Technology Conference of
  the NAACL, Companion Volume: Short Papers}, NAACL-Short '06, page 57–60,
  USA. Association for Computational Linguistics.

\bibitem[{Kumar et~al.(2020)Kumar, Bhotia, Kumar, and
  Chakraborty}]{kumar2020nurse}
Vaibhav Kumar, Tenzin~Singhay Bhotia, Vaibhav Kumar, and Tanmoy Chakraborty.
  2020.
\newblock \href {http://arxiv.org/abs/2006.01938} {Nurse is closer to woman
  than surgeon? mitigating gender-biased proximities in word embeddings}.

\bibitem[{Kurita et~al.(2019)Kurita, Vyas, Pareek, Black, and
  Tsvetkov}]{kurita2019measuring}
Keita Kurita, Nidhi Vyas, Ayush Pareek, Alan~W Black, and Yulia Tsvetkov. 2019.
\newblock Measuring bias in contextualized word representations.
\newblock \emph{arXiv preprint arXiv:1906.07337}.

\bibitem[{Lan et~al.(2020)Lan, Chen, Goodman, Gimpel, Sharma, and
  Soricut}]{lan2020albert}
Zhenzhong Lan, Mingda Chen, Sebastian Goodman, Kevin Gimpel, Piyush Sharma, and
  Radu Soricut. 2020.
\newblock \href {http://arxiv.org/abs/1909.11942} {Albert: A lite bert for
  self-supervised learning of language representations}.

\bibitem[{Liang et~al.(2020)Liang, Li, Zheng, Lim, Salakhutdinov, and
  Morency}]{liang2020towards}
Paul~Pu Liang, Irene~Mengze Li, Emily Zheng, Yao~Chong Lim, Ruslan
  Salakhutdinov, and Louis-Philippe Morency. 2020.
\newblock Towards debiasing sentence representations.
\newblock \emph{arXiv preprint arXiv:2007.08100}.

\bibitem[{Van~der Maaten and Hinton(2008)}]{van2008visualizing}
Laurens Van~der Maaten and Geoffrey Hinton. 2008.
\newblock Visualizing data using t-sne.
\newblock \emph{Journal of machine learning research}, 9(11).

\bibitem[{May et~al.(2019)May, Wang, Bordia, Bowman, and
  Rudinger}]{may2019measuring}
Chandler May, Alex Wang, Shikha Bordia, Samuel~R Bowman, and Rachel Rudinger.
  2019.
\newblock On measuring social biases in sentence encoders.
\newblock \emph{arXiv preprint arXiv:1903.10561}.

\bibitem[{Mikolov et~al.(2013)Mikolov, Chen, Corrado, and
  Dean}]{mikolov2013efficient}
Tomas Mikolov, Kai Chen, Greg Corrado, and Jeffrey Dean. 2013.
\newblock Efficient estimation of word representations in vector space.
\newblock \emph{arXiv preprint arXiv:1301.3781}.

\bibitem[{Nangia et~al.(2020)Nangia, Vania, Bhalerao, and
  Bowman}]{nangia-etal-2020-crows}
Nikita Nangia, Clara Vania, Rasika Bhalerao, and Samuel~R. Bowman. 2020.
\newblock \href {https://doi.org/10.18653/v1/2020.emnlp-main.154}
  {{C}row{S}-pairs: A challenge dataset for measuring social biases in masked
  language models}.
\newblock In \emph{Proceedings of the 2020 Conference on Empirical Methods in
  Natural Language Processing (EMNLP)}, pages 1953--1967, Online. Association
  for Computational Linguistics.

\bibitem[{Pennington et~al.(2014)Pennington, Socher, and
  Manning}]{pennington-etal-2014-glove}
Jeffrey Pennington, Richard Socher, and Christopher Manning. 2014.
\newblock \href {https://doi.org/10.3115/v1/D14-1162} {{G}love: Global vectors
  for word representation}.
\newblock In \emph{Proceedings of the 2014 Conference on Empirical Methods in
  Natural Language Processing ({EMNLP})}, pages 1532--1543, Doha, Qatar.
  Association for Computational Linguistics.

\bibitem[{Peters et~al.(2018)Peters, Neumann, Iyyer, Gardner, Clark, Lee, and
  Zettlemoyer}]{peters1802deep}
ME~Peters, M~Neumann, M~Iyyer, M~Gardner, C~Clark, K~Lee, and L~Zettlemoyer.
  2018.
\newblock Deep contextualized word representations. arxiv 2018.
\newblock \emph{arXiv preprint arXiv:1802.05365}.

\bibitem[{Rudinger et~al.(2018)Rudinger, Naradowsky, Leonard, and {Van
  Durme}}]{rudinger-EtAl:2018:N18}
Rachel Rudinger, Jason Naradowsky, Brian Leonard, and Benjamin {Van Durme}.
  2018.
\newblock Gender bias in coreference resolution.
\newblock In \emph{Proceedings of the 2018 Conference of the North American
  Chapter of the Association for Computational Linguistics: Human Language
  Technologies}, New Orleans, Louisiana. Association for Computational
  Linguistics.

\bibitem[{Socher et~al.(2013)Socher, Perelygin, Wu, Chuang, Manning, Ng, and
  Potts}]{socher2013recursive}
Richard Socher, Alex Perelygin, Jean Wu, Jason Chuang, Christopher~D Manning,
  Andrew~Y Ng, and Christopher Potts. 2013.
\newblock Recursive deep models for semantic compositionality over a sentiment
  treebank.
\newblock In \emph{Proceedings of the 2013 conference on empirical methods in
  natural language processing}, pages 1631--1642.

\bibitem[{Wang et~al.(2018)Wang, Singh, Michael, Hill, Levy, and
  Bowman}]{wang2018glue}
Alex Wang, Amanpreet Singh, Julian Michael, Felix Hill, Omer Levy, and Samuel~R
  Bowman. 2018.
\newblock Glue: A multi-task benchmark and analysis platform for natural
  language understanding.
\newblock \emph{arXiv preprint arXiv:1804.07461}.

\bibitem[{Wang et~al.(2020)Wang, Lin, Rajani, McCann, Ordonez, and
  Xiong}]{wang2020double}
Tianlu Wang, Xi~Victoria Lin, Nazneen~Fatema Rajani, Bryan McCann, Vicente
  Ordonez, and Caiming Xiong. 2020.
\newblock Double-hard debias: Tailoring word embeddings for gender bias
  mitigation.
\newblock \emph{arXiv preprint arXiv:2005.00965}.

\bibitem[{Warstadt et~al.(2019)Warstadt, Singh, and
  Bowman}]{warstadt2019neural}
Alex Warstadt, Amanpreet Singh, and Samuel~R Bowman. 2019.
\newblock Neural network acceptability judgments.
\newblock \emph{Transactions of the Association for Computational Linguistics},
  7:625--641.

\bibitem[{Webster et~al.(2020)Webster, Wang, Tenney, Beutel, Pitler, Pavlick,
  Chen, and Petrov}]{webster2020measuring}
Kellie Webster, Xuezhi Wang, Ian Tenney, Alex Beutel, Emily Pitler, Ellie
  Pavlick, Jilin Chen, and Slav Petrov. 2020.
\newblock Measuring and reducing gendered correlations in pre-trained models.
\newblock \emph{arXiv preprint arXiv:2010.06032}.

\bibitem[{Zhao et~al.(2019)Zhao, Wang, Yatskar, Cotterell, Ordonez, and
  Chang}]{zhao2019gender}
Jieyu Zhao, Tianlu Wang, Mark Yatskar, Ryan Cotterell, Vicente Ordonez, and
  Kai-Wei Chang. 2019.
\newblock Gender bias in contextualized word embeddings.
\newblock \emph{arXiv preprint arXiv:1904.03310}.

\bibitem[{Zhao et~al.(2018{\natexlab{a}})Zhao, Wang, Yatskar, Ordonez, and
  Chang}]{zhao2018gender}
Jieyu Zhao, Tianlu Wang, Mark Yatskar, Vicente Ordonez, and Kai-Wei Chang.
  2018{\natexlab{a}}.
\newblock Gender bias in coreference resolution: Evaluation and debiasing
  methods.
\newblock \emph{arXiv preprint arXiv:1804.06876}.

\bibitem[{Zhao et~al.(2018{\natexlab{b}})Zhao, Zhou, Li, Wang, and
  Chang}]{zhao2018learning}
Jieyu Zhao, Yichao Zhou, Zeyu Li, Wei Wang, and Kai-Wei Chang.
  2018{\natexlab{b}}.
\newblock Learning gender-neutral word embeddings.
\newblock \emph{arXiv preprint arXiv:1809.01496}.

\bibitem[{Zhu et~al.(2015)Zhu, Kiros, Zemel, Salakhutdinov, Urtasun, Torralba,
  and Fidler}]{Zhu_2015_ICCV}
Yukun Zhu, Ryan Kiros, Rich Zemel, Ruslan Salakhutdinov, Raquel Urtasun,
  Antonio Torralba, and Sanja Fidler. 2015.
\newblock Aligning books and movies: Towards story-like visual explanations by
  watching movies and reading books.
\newblock In \emph{The IEEE International Conference on Computer Vision
  (ICCV)}.

\end{thebibliography}
\bibliographystyle{acl_natbib}

\appendix

\section{Appendices}
\label{sec:appendix}
\subsection{Related Work}
\subsubsection{Contextual Embeddings}
\label{subsec:contextualEmbeddingRelatedWorks}
Existing works mainly focus on debiasing static word embeddings. The works related to contextual embeddings are mainly for analyzing gender bias in contextualized architectures.  \citet{may2019measuring} propose a more generalized approach for measuring bias in contextualized word representations called SEAT (Sentence Encoder Association Test). Also, \citet{kurita2019measuring} provide a template-based approach to quantify bias. \citet{nangia-etal-2020-crows} and \citet{kurita2019measuring} both explore various techniques to understand the bias propagated by large language models like BERT and use the log likelihood score for each masked token as an indicator to analyze the bias contained in the model.

On the other hand, few recent research tried to tackle the contextual aspect of bias prevalent in word representations. Mostly, data-augmentation strategies like CDA (counterfactual data augmentation) have become popular, which augment training data using controlled perturbations to names or demographic attributes. \citet{zhao2019gender} looks into debiasing contextual word embeddings like ELMo \citep{peters1802deep} by using simple yet effective pre-processing and post-processing techniques. The proposed solution of data augmentation is performed by replacing gender-related entities in the dataset with words of the opposite gender and then training on the union of the original data and this swapped data. \citet{webster2020measuring} follows a similar approach that helps mitigate gender bias with minimal trade-offs for large-scale language models. Though these works address the problem of biases stemming from data imbalances by associating every gender neutral word to both male and female context equally, they do not discuss the efficacy of this technique for evaluating representations quality for unpaired gendered words like beard or bikini etc which will be further discussed in sec.~\ref{subsec:bert}.

\citet{zhao2019gender} provides empirical results of reduced bias when averaging the ELMo representations of words in different gendered contexts. \citet{webster2020measuring} observed that increasing the dropout rates for both BERT \citep{devlin-etal-2019-bert} and ALBERT \citep{lan2020albert} reduces bias with little loss in accuracy. 

\subsubsection{Static (Non-contextual) Embeddings}
\label{subsec:staticEmbeddingRelatedWorks}
\citet{bolukbasi2016man} was the first to introduce a post-hoc debiasing technique that aims to ensure that gender-neutral words are of equal distance from gender-specific words like man/woman. They introduce the concept of a gender subspace which exists in all word embeddings by computing the principal components across the difference between 10 pairs of the most frequently occurring gender paired words as shown in Figure \ref{wordpair}(a) (in Appendix \ref{appfigure}). Once we estimate the principal components for the 10 pairs of vectors as shown in Figure \ref{wordpair}(b)(in Appendix \ref{appfigure}), we find that the first principal component is most informative and hence is most likely to represent the gender subspace. Two approaches are discussed, Neutralize-Equalize and Soften, which help reduce the projection of these representations in the gender subspace.
% Then, we proceed to remove the bias captured by the embeddings in two approaches: \emph{Neutralize and Equalize} and \emph{Soften}.
    
% As for \emph{Neutralize and Equalize}, they first \emph{neutralize} the words such that their representations have zero projection in the gender subspace. Then, they \emph{equalize} the words by enforcing all gender-neutral words to be equidistant from all gendered words in the equality set. \emph{Equalize} is necessary for preserving the equivalent analogies, i.e. \textit{father:male::mother:female}. However, certain distinctions might be lost in gender-related terminology like \textit{"grandfather a regulation"} v/s \textit{"grandmother a regulation"}. On the other hand, \emph{Soften} preserves pairwise inner products among all word vectors while limiting the projection of gender-neutral words on the gender direction by applying a linear transformation to word vectors.
 %\textbf{Gender neutral embeddings}: In this in-processing technique, 
\citet{zhao2018learning} propose a  training curriculum that aims to restrict the gender influence of embeddings to certain dimensions and keep the rest of the dimensions free of gender influence. During inference, they simply remove the dimensions in which the gender are encoded and use the learned \textit{gender-neutral} representations for downstream tasks. 
% They showed that this approach gave much more neutral embeddings compared to the approach introduced in \citet{bolukbasi2016man}.
% \textbf{Double Hard Debiasing}: 

The recent work from \citet{wang2020double}, double-hard debiasing, extends the work from \citet{bolukbasi2016man} by post-processing the embeddings with an additional debiasing step after the initial hard-debiased representations. The authors observe that semi-agnostic corpus tendencies such as word frequencies tamper the gender neutrality of most word representations. They prune this particular statistical dependency from word embeddings and further show that this marginally improves the gender bias mitigation while maintaining the distributional semantics of the embedding representation. 
% It uses the same approach as \citet{bolukbasi2016man} to identify the word frequency subspace and neutralize it from the already hard-debiased word representation.
This approach was shown to perform marginally better on datasets like OntoNotes \citep{10.5555/1614049.1614064} and WinoBias \citep{zhao2018gender}.
   
Another recent work from \citet{kumar2020nurse} proposes a brand-new algorithm, Repulsion, Attraction, and Neutralization based Debiasing (RANDebias) which not only eliminates the bias present in a word vector but also alters the spatial distribution of its neighboring vectors, achieving a bias-free setting while maintaining minimal semantic offset. 
% Most of the existing debiasing methods are primarily concerned with debiasing a word vector by minimising its projection on the gender direction. Therefore, they 
The authors argue that current work tends to ignore the relationship between a gender-neutral word vector and its neighbors, thus failing to remove the gender bias encoded as illicit proximities between words. By mitigating both direct and gender-based proximity bias while adding minimal impact to the semantic and analogical properties of the word embedding, the authors (with the help of a newly proposed metric (GIPE)) help substantiate the claims of removing bias.
\clearpage
\subsection{Quantitative Analysis: Static (Non-contextual) Embeddings}
\subsubsection{Clustering Labels for NER-M + EGE embeddings}
\label{sec:appendix:clusterList}
\begin{table}[h!]
\small
\begin{tabular}{|c|ccc|}
\hline
\textbf{Cluster 1}                      & \multicolumn{3}{c|}{\textbf{Cluster 2}}                                                                                         \\ \hline
{\color[HTML]{FE0000} trucker(M)}       & {\color[HTML]{FE0000} legislator(M)}     & {\color[HTML]{036400} performer(F)}     & {\color[HTML]{036400} homemaker(F)}        \\
{\color[HTML]{FE0000} mobster(M)}       & {\color[HTML]{036400} dancer(F)}         & {\color[HTML]{036400} teacher(F)}       & {\color[HTML]{036400} stylist(F)}          \\
{\color[HTML]{FE0000} custodian(M)}     & {\color[HTML]{036400} undersecretary(F)} & {\color[HTML]{FE0000} farmer(M)}        & {\color[HTML]{FE0000} mathematician(M)}    \\
{\color[HTML]{FE0000} ballplayer(M)}    & {\color[HTML]{036400} maid(F)}           & {\color[HTML]{FE0000} minister(M)}      & {\color[HTML]{036400} clerk(F)}            \\
{\color[HTML]{036400} artiste(F)}       & {\color[HTML]{FE0000} superintendent(M)} & {\color[HTML]{036400} secretary(F)}     & {\color[HTML]{036400} educator(F)}         \\
{\color[HTML]{036400} therapist(F)}     & {\color[HTML]{FE0000} commander(M)}      & {\color[HTML]{036400} violinist(F)}     & {\color[HTML]{036400} environmentalist(F)} \\
{\color[HTML]{036400} bookkeeper(F)}    & {\color[HTML]{FE0000} sergeant(M)}       & {\color[HTML]{FE0000} colonel(M)}       & {\color[HTML]{036400} vocalist(F)}         \\
{\color[HTML]{036400} hairdresser(F)}   & {\color[HTML]{036400} caretaker(F)}      & {\color[HTML]{FE0000} drummer(M)}       & {\color[HTML]{036400} hooker(F)}           \\
{\color[HTML]{036400} radiologist(F)}   & {\color[HTML]{036400} lyricist(F)}       & {\color[HTML]{036400} choreographer(F)} & {\color[HTML]{FE0000} tycoon(M)}           \\
{\color[HTML]{FE0000} neurosurgeon(M)}  & {\color[HTML]{036400} librarian(F)}      & {\color[HTML]{036400} ballerina(F)}     & {\color[HTML]{FE0000} magician(M)}         \\
{\color[HTML]{036400} pediatrician(F)}  & {\color[HTML]{FE0000} pastor(M)}         & {\color[HTML]{036400} mediator(F)}      & {\color[HTML]{FE0000} carpenter(M)}        \\
{\color[HTML]{FE0000} plumber(M)}       & {\color[HTML]{FE0000} warrior(M)}        & {\color[HTML]{FE0000} laborer(M)}       & {\color[HTML]{FE0000} coach(M)}            \\
{\color[HTML]{036400} nanny(F)}         & {\color[HTML]{FE0000} philosopher(M)}    & {\color[HTML]{036400} naturalist(F)}    & {\color[HTML]{036400} artist(F)}           \\
{\color[HTML]{036400} paralegal(F)}     & {\color[HTML]{FE0000} marshal(M)}        & {\color[HTML]{FE0000} preacher(M)}      & {\color[HTML]{036400} soloist(F)}          \\
{\color[HTML]{FE0000} electrician(M)}   & {\color[HTML]{FE0000} senator(M)}        & {\color[HTML]{FE0000} bodyguard(M)}     & {\color[HTML]{036400} instructor(F)}       \\
{\color[HTML]{036400} housekeeper(F)}   & {\color[HTML]{FE0000} physicist(M)}      & {\color[HTML]{FE0000} astronaut(M)}     & {\color[HTML]{036400} psychiatrist(F)}     \\
{\color[HTML]{036400} receptionist(F)}  & {\color[HTML]{FE0000} warden(M)}         & {\color[HTML]{FE0000} magistrate(M)}    & {\color[HTML]{FE0000} sailor(M)}           \\
{\color[HTML]{FE0000} lawmaker(M)}      & {\color[HTML]{FE0000} sportswriter(M)}   & {\color[HTML]{FE0000} boxer(M)}         & {\color[HTML]{FE0000} surgeon(M)}          \\
{\color[HTML]{FE0000} butcher(M)}       & {\color[HTML]{FE0000} ranger(M)}         & {\color[HTML]{036400} publicist(F)}     & {\color[HTML]{FE0000} assassin(M)}         \\
{\color[HTML]{036400} dermatologist(F)} & {\color[HTML]{036400} organist(F)}       & {\color[HTML]{036400} socialite(F)}     & {\color[HTML]{FE0000} president(M)}        \\
{\color[HTML]{036400} realtor(F)}       & {\color[HTML]{FE0000} commissioner(M)}   & {\color[HTML]{036400} aide(F)}          & {\color[HTML]{036400} nurse(F)}            \\
{\color[HTML]{FE0000} skipper(M)}       & {\color[HTML]{FE0000} officer(M)}        & {\color[HTML]{036400} tutor(F)}         & {\color[HTML]{036400} treasurer(F)}        \\
{\color[HTML]{FE0000} janitor(M)}       & {\color[HTML]{FE0000} lieutenant(M)}     & {\color[HTML]{036400} planner(F)}       &                                            \\ \hline
\end{tabular}
\caption{Table showing how clusters are grouped for our \textbf{NER-M + EGE} debiasing embeddings. The following cluster does not show signs of conforming to stereotypical clusters influenced by gender bias.}
\end{table}

\clearpage
\subsection{Qualitative Analysis: Static (Non-contextual) Embeddings}
\label{appendix:hairdresser_nanny}
\subsubsection{Word Cloud Visualization for `nanny' and `hairdresser'}
\begin{figure}[h!]
\includegraphics[scale=0.49]{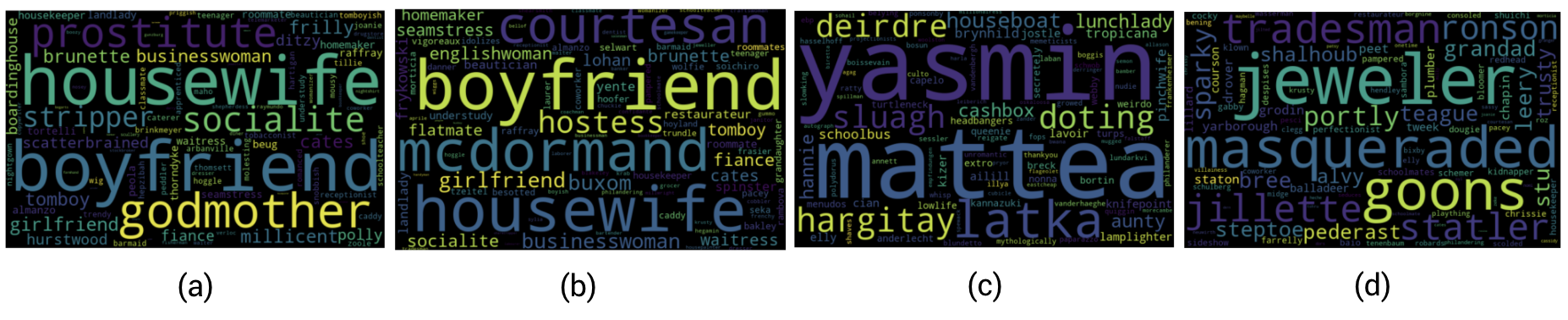}
\label{fig:hairdresser}
% \end{figure}
% \begin{figure}[h!]
\includegraphics[scale=0.286]{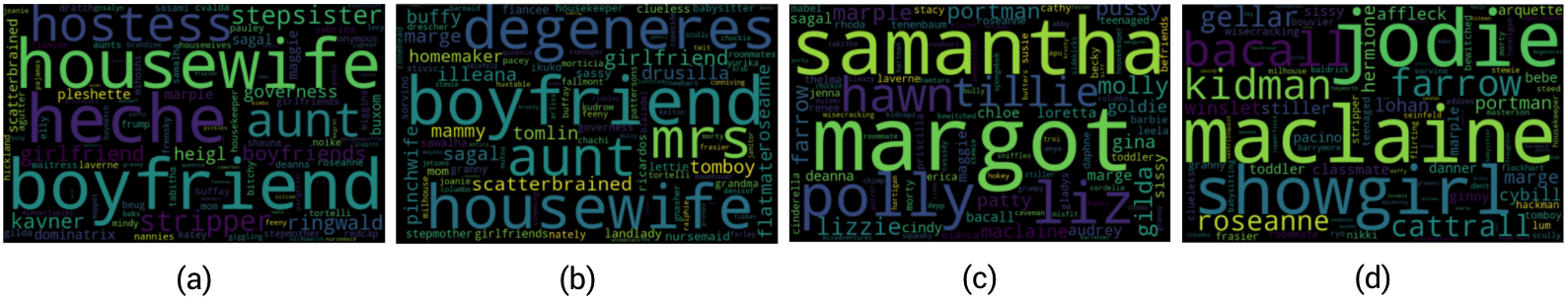}
\caption{The word cloud visualizations for the neighbours of the word `hairdresser' (top) and `nanny' (bottom) (size coded by bias by projection) for (a) \textbf{original Word2Vec}, (b) \textbf{NER-M Word2Vec}, (c) \textbf{EGE Word2Vec}, and (d) \textbf{NER-M + EGE Word2Vec}. We observe that EGE in itself is sufficient to reduce the proximity bias of gendered words, however this results in nanny clustering close to stereotypically female names. This phenomenon persists even after removing the most common names but however does result in a much lower direct and proximity bias values. A detailed analysis for neighbours of `nanny' is provided in Appendix \ref{sec:appendix:nanny}}
\label{fig:nanny}
\end{figure}

\clearpage

\subsubsection{Neighbour Analysis for `nanny'}
\label{sec:appendix:nanny}

\begin{table}[h!]
\centering
\small
\begin{tabular}{|c|c|c|c|c|c|c|c|c|}
\hline
\multicolumn{9}{|c|}{\textbf{Neighbour Analysis for `nanny'}}                                                                                                    \\ \hline
\textbf{Embedding}      & \multicolumn{2}{c|}{Original} & \multicolumn{2}{c|}{NER-M}  & \multicolumn{2}{c|}{EGE}          & \multicolumn{2}{c|}{NER-M + EGE}     \\ \hline
\textbf{Direct Bias}    & \multicolumn{2}{c|}{0.3059}   & \multicolumn{2}{c|}{0.2919} & \multicolumn{2}{c|}{0.3308}       & \multicolumn{2}{c|}{\textbf{0.2259}} \\ \hline
\textbf{Proximity Bias} & \multicolumn{2}{c|}{0.65}     & \multicolumn{2}{c|}{0.38}   & \multicolumn{2}{c|}{\textbf{0.0}} & \multicolumn{2}{c|}{\textbf{0.0}}    \\ \hline
\multicolumn{9}{|c|}{\textbf{Neighbour /Bias by projection}}                                                                                                     \\ \hline
0                       & boyfriend        & 0.386991   & boyfriend      & 0.400996   & polly            & 0.282503       & bacall             & 0.201971        \\ \hline
1                       & housewife        & 0.368933   & housewife      & 0.327343   & tillie           & 0.247825       & farrow             & 0.197205        \\ \hline
2                       & aunt             & 0.345754   & aunt           & 0.272739   & lizzie           & 0.244246       & gellar             & 0.185702        \\ \hline
3                       & governess        & 0.302615   & roseanne       & 0.219477   & gilda            & 0.242196       & roseanne           & 0.183616        \\ \hline
4                       & pleshette        & 0.296761   & flatmate       & 0.214433   & marple           & 0.241026       & marge              & 0.164832        \\ \hline
5                       & scatterbrained   & 0.279132   & fiancee        & 0.190719   & sagal            & 0.208192       & cybill             & 0.160163        \\ \hline
6                       & girlfriends      & 0.267669   & housekeeper    & 0.185169   & maggie           & 0.207090       & marple             & 0.126461        \\ \hline
7                       & katey            & 0.240061   & tortelli       & 0.179968   & chloe            & 0.206370       & frasier            & 0.109756        \\ \hline
8                       & roseanne         & 0.237083   & roommates      & 0.179795   & deanna           & 0.195022       & nikki              & 0.106050        \\ \hline
9                       & housekeeper      & 0.236778   & grandma        & 0.179707   & teenaged         & 0.194433       & clueless           & 0.092178        \\ \hline
10                      & waitress         & 0.227350   & kudrow         & 0.176179   & laverne          & 0.185224       & bewitched          & 0.091465        \\ \hline
11                      & mom              & 0.208003   & chachi         & 0.154935   & jenna            & 0.181688       & morty              & 0.076856        \\ \hline
12                      & fiancee          & 0.205739   & pacey          & 0.146833   & roseanne         & 0.181319       & housekeeper        & 0.071950        \\ \hline
13                      & baywatch         & 0.178570   & frasier        & 0.126824   & morty            & 0.149673       & stewie             & 0.068744        \\ \hline
14                      & morgendorffer    & 0.157481   & kelton         & 0.115218   & abby             & 0.141383       & columbo            & 0.050975        \\ \hline
15                      & jughead          & 0.152763   & columbo        & 0.100723   & apu              & 0.122025       & babysitter         & 0.042014        \\ \hline
16                      & queenie          & 0.147512   & boomhauer      & 0.092048   & mulder           & 0.102394       & copperfield        & 0.039764        \\ \hline
17                      & magrat           & 0.147462   & conniving      & 0.085128   & granny           & 0.096720       & telly              & 0.031565        \\ \hline
18                      & birdsworth       & 0.129046   & rickles        & 0.079178   & timmy            & 0.061670       & janitor            & 0.012138        \\ \hline
19                      & grandpa          & 0.019874   & grandpa        & 0.047922   & bartender        & 0.058724       & winchell           & 0.006846        \\ \hline

\end{tabular}
\caption{We analyze the Direct Bias (Bias by projection on the PCA based gender direction) and Proximity Bias (Ratio of biased neighbours by Indirect Bias) for various methods proposed. We also analyze the neighbours for each method. We see that once we explicity code the gender, the neighbours for `nanny' separate themselves from female gendered words (resulting in a proximity bias of 0) and cluster closely with stereotypically female names. Although our NER-M + EGE is not completely effective in weeding out all names, we find that with the amount pruned, terms like babysitter (15) do appear in the closest neighbours for the word nanny which are more semantically related than the neighbours for other methods. We also see a stark decrease in the Direct and Proximity bias validating the success of our method.}
\end{table}

\clearpage
\subsubsection{Gender Clustering of baselines and proposed approaches}
\label{appendix:genderCluster}
\begin{figure}[h!]
\centering
\includegraphics[scale=0.58]{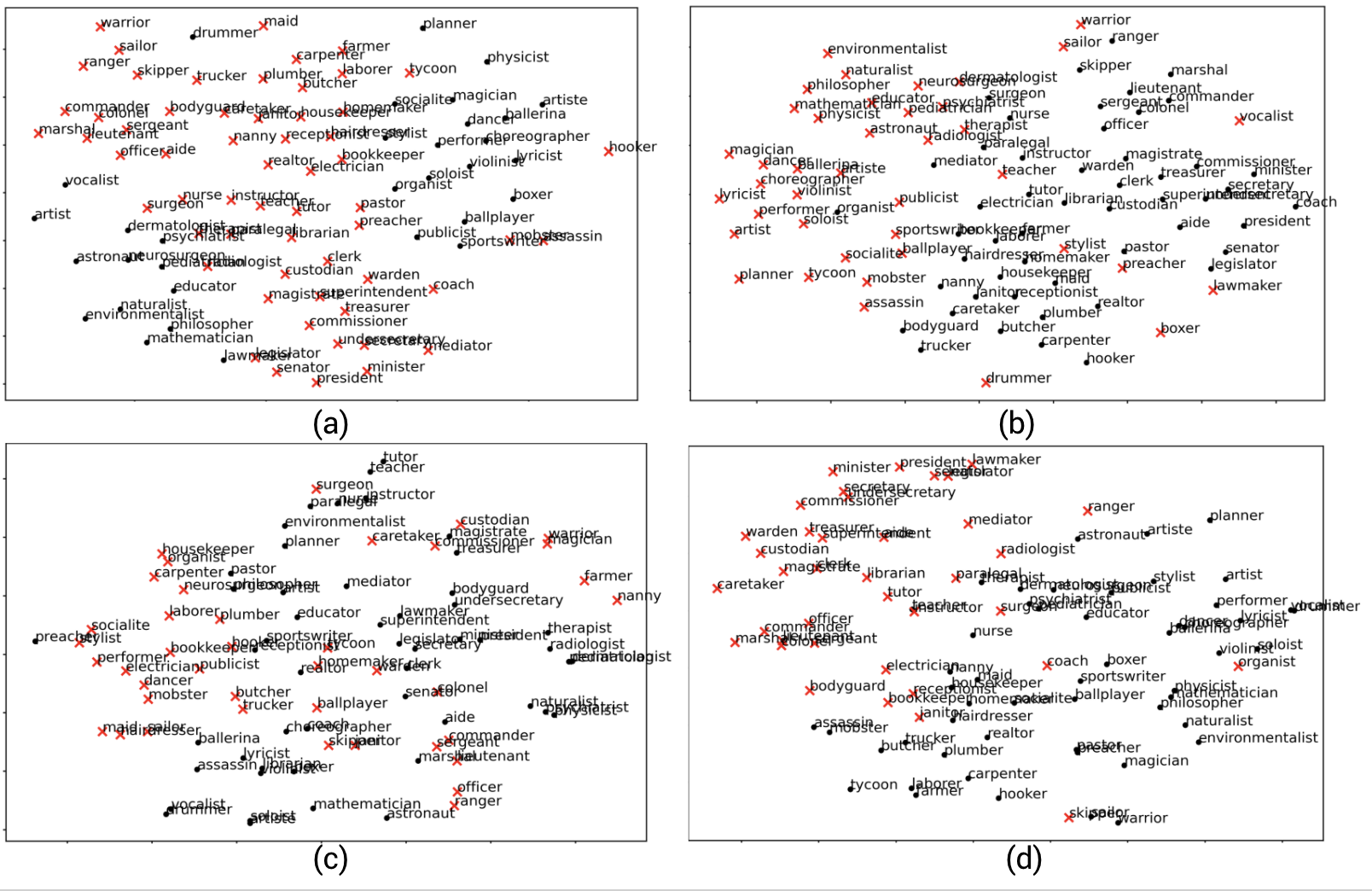}
\caption{TSNE visualizations of the vectors during gender clustering for (a) \textbf{Hard Debias} \citep{bolukbasi2016man}, (b) \textbf{Double Hard Debias} \citep{wang2020double}, (c) \textbf{RAN Debias} \citep{kumar2020nurse}, and (d) \textbf{NER-M} (Ours). From manual inspection, it is clear that each method retains some notion of smenatics in the embeddings, the clusters are not as tightly coupled as seen in (b) Figure \ref{fig:cluster1}. The semantic effectiveness of our approach is also verified by the near-perfect score we achieve in resisting gender-stereotypical clustering. (See Table \ref{table:clusterAccuracy})}
\label{fig:cluster2}
\end{figure}
\clearpage
\subsubsection{Neighbour Analysis for `nurse'}
\begin{figure}[h!]
\centering
\includegraphics[scale=0.52]{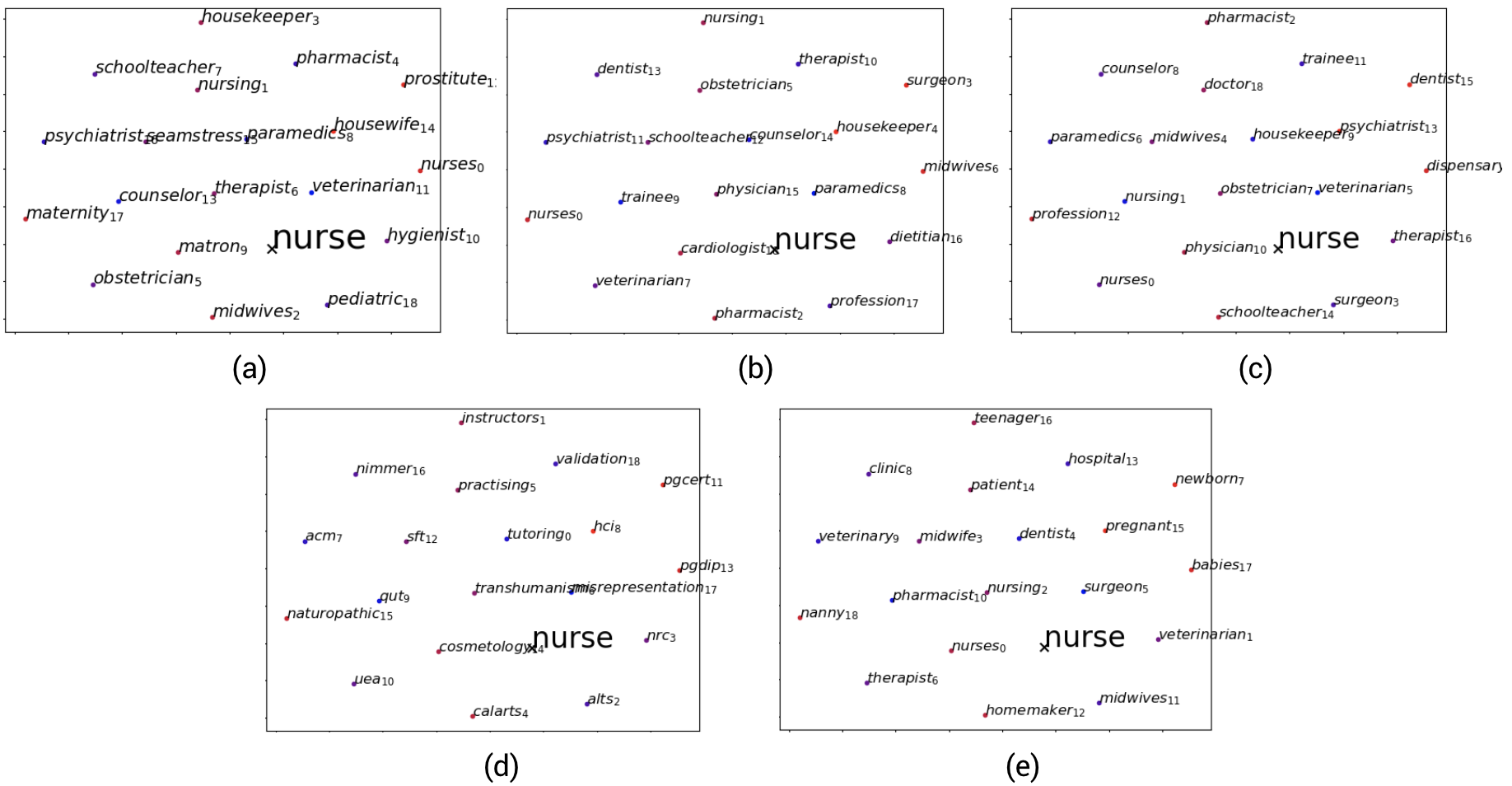}
\caption{ TSNE plot for neighbours of nurse (color coded by bias by projection) for (a) \textbf{Original Word2Vec} \citep{mikolov2013efficient}, (b) \textbf{Hard Debias} \citep{bolukbasi2016man}, (c) \textbf{Double Hard Debias}\citep{wang2020double}, (d) \textbf{RAN Debias}\citep{kumar2020nurse}, and (e) \textbf{NER-M + EGE} (Ours). In (a), we observe words like prostitute, seamstress and schoolteacher appear in neighbours for nurse which is a clear case of bias via occupational stereotypes. In (b) and (c) we see that the semantics of the neighbours are better than the original, but they still exhibit traces of bias via words like schoolteacher and housekeeper. (d) seems to completely veer away from the related words from a semantic viewpoint and we hypothesize this may be due to the step where they neutralize the frequency distribution of biased words. Our method (e), showcases much more semantically related neighbours where we see words like clinic, hospital and patient appear in the top 20 neighbours.}
\label{fig:nursebig}
\end{figure}

\clearpage
\subsubsection{Visualization of Cluster shift}
\label{appendix:less_annotate}

\begin{figure}[h!]
\centering
\includegraphics[scale=0.47]{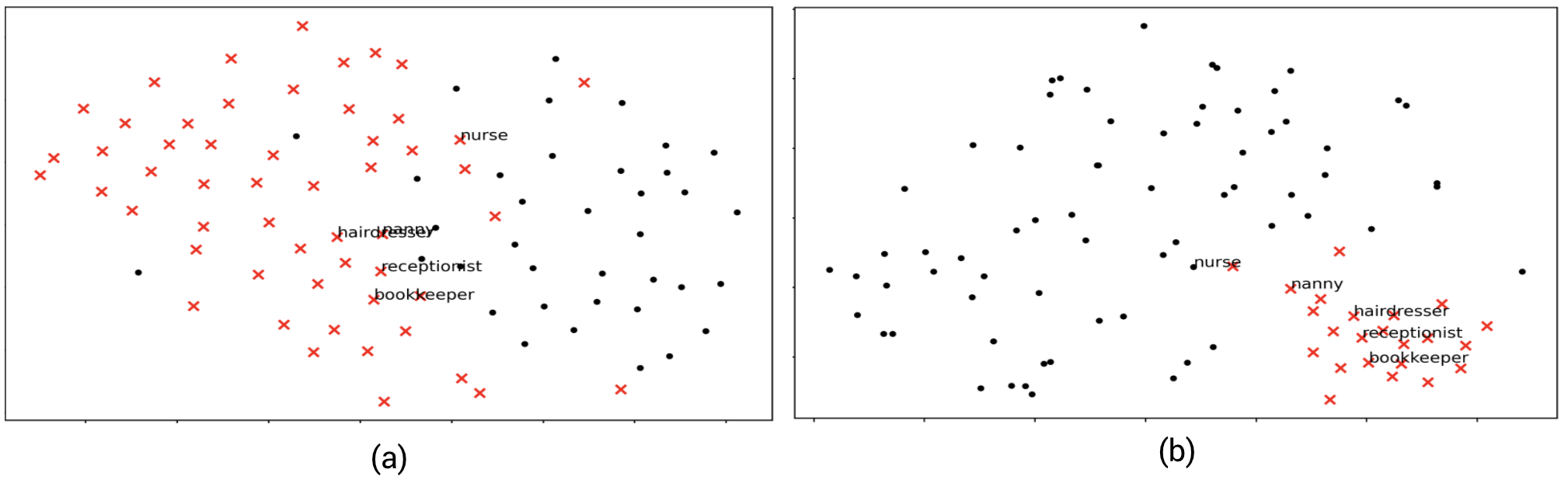}
\caption{Gender clustering of (a) \textbf{Original}, (b) \textbf{NER-M + EGE} embeddings. Notice the change in clusters for nurse and the increase in separation between hairdresser and nanny.}
\label{fig:less}
\end{figure}

\clearpage
\subsubsection{Neighbour analysis for `molly' }
\label{sec:appendix:molly}
    
\begin{table}[h!]
\begin{center}
\begin{tabular}{|c|c|c|c|c|}
\hline
\multicolumn{5}{|c|}{\textbf{Neighbour Analysis for `molly'}}                               \\ \hline
\textbf{Embedding}      & \multicolumn{2}{c|}{Original} & \multicolumn{2}{c|}{EGE}          \\ \hline
\textbf{Direct Bias}    & \multicolumn{2}{c|}{0.2703}   & \multicolumn{2}{c|}{0.2460}       \\ \hline
\textbf{Proximity Bias} & \multicolumn{2}{c|}{0.73}     & \multicolumn{2}{c|}{\textbf{0.0}} \\ \hline
\multicolumn{5}{|c|}{\textbf{Neighbour /Bias by projection}}                                \\ \hline
0                       & housewife      & 0.368933     & carrie           & 0.323712       \\ \hline
1                       & sally          & 0.335042     & heather          & 0.284811       \\ \hline
2                       & suffragist     & 0.320435     & polly            & 0.282503       \\ \hline
3                       & loretta        & 0.310904     & arquette         & 0.281774       \\ \hline
4                       & millicent      & 0.309910     & jill             & 0.265960       \\ \hline
5                       & ringwald       & 0.305847     & linda            & 0.257541       \\ \hline
6                       & marsha         & 0.286048     & hawn             & 0.249712       \\ \hline
7                       & maggie         & 0.279420     & debbie           & 0.247345       \\ \hline
8                       & picon          & 0.268772     & brenda           & 0.237183       \\ \hline
9                       & gebbie         & 0.268752     & jenny            & 0.219900       \\ \hline
\end{tabular}
\caption{The table highlights how the proposed EGE method while succeeding in eliminating the Proximity Bias (Ratio of biased neighbours by Indirect Bias) still clusters along with stereotypically female names. While occupations like housewife and suffragist are removed from the neighbours for `molly', the fact remains that the implicit clustering of female names together results in transitive transmission of bias between female stereotypical professions.}
\end{center}
\end{table}

\clearpage
\subsection{Limitations}
\subsubsection{Neighbour Word Cloud Visualization of `dirty'}
\label{appendix:dirty}
\begin{figure}[h!]
\centering
\includegraphics[scale=0.53]{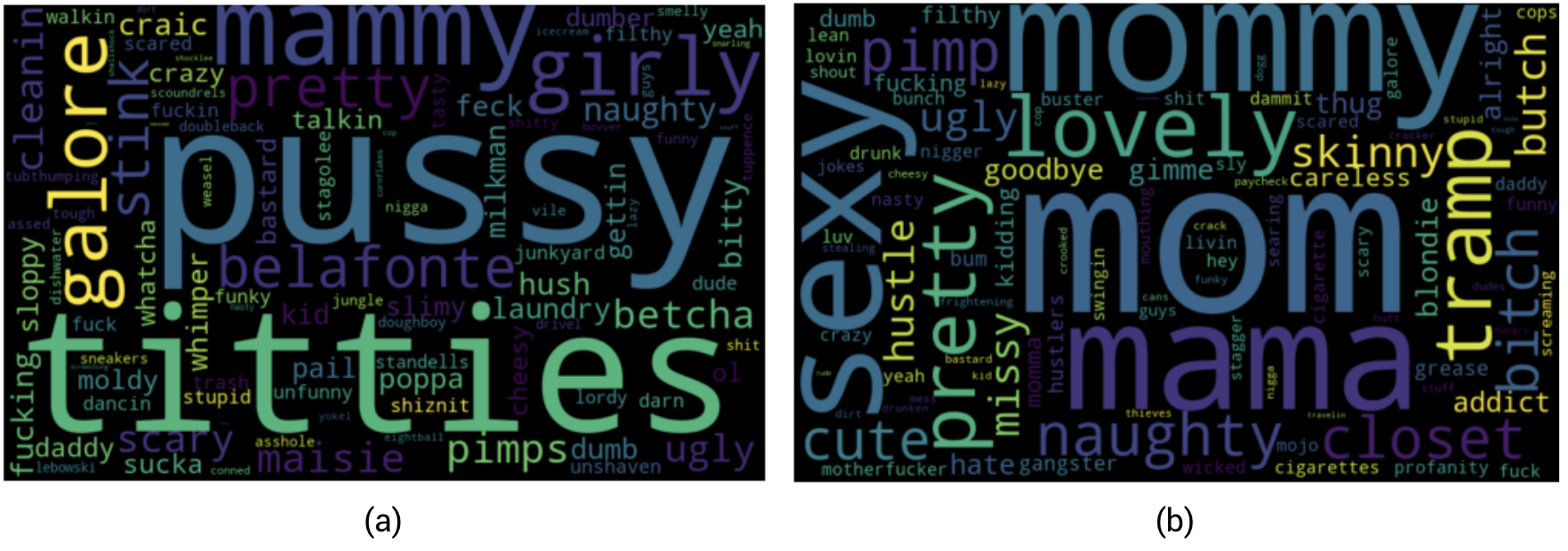}
\caption{The word cloud visualizations for the word `dirty' for (a) \textbf{Original}, (b) \textbf{NER-M + EGE} embeddings (size coded by bias by projection). While our technique does attempt to make the associations less profane or sexually explicit, it fails to eliminate the association to female stereotypical words. This opens up to the possibility that adjectives perhaps collect bias from a different secondary source (akin to how professions imbibe them via names).}
\label{fig:dirty}
\end{figure}

\end{document}